\documentclass[lettersize,journal]{IEEEtran}
\usepackage{amsmath,amsfonts}
\usepackage{algorithmic}
\usepackage{algorithm}
\usepackage{array}
\usepackage[caption=false,font=normalsize,labelfont=sf,textfont=sf]{subfig}
\usepackage{textcomp}
\usepackage{stfloats}
\usepackage{url}
\usepackage{verbatim}
\usepackage{graphicx}
\usepackage{booktabs}
\usepackage{multirow}
\usepackage{makecell}
\usepackage{hyperref} 
\usepackage[nocompress]{cite}
\begin{document}

\title{Underlying Semantic Diffusion for Effective and Efficient In-Context Learning}
\author{Zhong Ji,~\IEEEmembership{Senior Member,~IEEE,} 
        Weilong Cao, 
        Yan Zhang, 
        Yanwei Pang,~\IEEEmembership{Senior Member,~IEEE,}\\
        Jungong Han,~\IEEEmembership{Senior Member,~IEEE,}
        Xuelong Li,~\IEEEmembership{Fellow,~IEEE}
\thanks{This work was supported by the National Natural Science Foundation of China (NSFC) under Grants 62441235 and 62176178. \emph{(Corresponding author: Yan Zhang.)}}
\thanks{Zhong Ji, Weilong Cao, Yan Zhang, and Yanwei Pang are with the School of Electrical and Information Engineering, Tianjin Key Laboratory of
Brain-inspired Intelligence Technology, Tianjin University, Tianjin 300072,
China (e-mail: jizhong@tju.edu.cn; caoweilong@tju.edu.cn; yzhang1995@tju.edu.cn; pyw@tju.edu.cn).

Jungong Han is with the Department of Automation, Tsinghua University, Beijing 100084, China (e-mail: jghan@tsinghua.edu.cn).

Xuelong Li is with the Institute of Artificial Intelligence (TeleAI), China Telecom Corp Ltd, 31 Jinrong Street, Beijing 100033, China (e-mail: xuelong li@ieee.org).
}}

\markboth{Journal of \LaTeX\ Class Files,~Vol.~14, No.~8, August~2021}%
{Shell \MakeLowercase{\textit{et al.}}: A Sample Article Using IEEEtran.cls for IEEE Journals}


\maketitle

\begin{abstract}
Diffusion models has emerged as a powerful framework for tasks like image controllable generation and dense prediction. However, existing models often struggle to capture underlying semantics (e.g., edges, textures, shapes) and effectively utilize in-context learning, limiting their contextual understanding and image generation quality. Additionally, high computational costs and slow inference speeds hinder their real-time applicability.
To address these challenges, we propose Underlying Semantic Diffusion (US-Diffusion), an enhanced diffusion model that boosts underlying semantics learning, computational efficiency, and in-context learning capabilities on multi-task scenarios. We introduce Separate \& Gather Adapter (SGA), which decouples input conditions for different tasks while sharing the architecture, enabling better in-context learning and generalization across diverse visual domains. We also present a Feedback-Aided Learning (FAL) framework, which leverages feedback signals to guide the model in capturing semantic details and dynamically adapting to task-specific contextual cues. Furthermore, we propose a plug-and-play Efficient Sampling Strategy (ESS) for dense sampling at time steps with high-noise levels, which aims at optimizing training and inference efficiency while maintaining strong in-context learning performance.
Experimental results demonstrate that US-Diffusion outperforms the state-of-the-art method, achieving an average reduction of 7.47 in FID on Map2Image tasks and an average reduction of 0.026 in RMSE on Image2Map tasks, while achieving approximately 9.45× faster inference speed. 
Our method also demonstrates superior training efficiency and in-context learning capabilities, excelling in new datasets and tasks, highlighting its robustness and adaptability across diverse visual domains. The source code will be released at \href{https://github.com/dragon-cao/US-Diffusion}{https://github.com/dragon-cao/US-Diffusion}.
\end{abstract}


\section{Introduction}
\IEEEPARstart{R}{ecent} advances in generative foundation models like GPT-4\cite{gpt} and DeepSeek-R1\cite{deepseek} revolutionize Natural Language Processing (NLP)\cite{nlp1,nlp2,nlp3}, inspiring similar breakthroughs in Computer Vision (CV)\cite{cv1,cv2,cv7,cv9}. Among these, diffusion models\cite{ddpm,diffusion1,diffusion2,diffusion3,stablediffusion} now emerge as powerful visual generators, gaining traction in image generation and becoming increasingly popular for CV tasks such as image controllable generation\cite{imagegeneration1,imagegeneration2,stablediffusion,controlnet} and dense prediction\cite{dense,semantic1,semantic2,depth1,depth2}. 
For instance, Prompt Diffusion (abbreviated as PromptDiff in this work for convenience) \cite{promptdiffusion} preliminarily unlocks the in-context learning ability of diffusion models, achieving impressive results across various visual benchmarks.
\begin{figure*}[!t]
\centering
\includegraphics[scale=0.965]{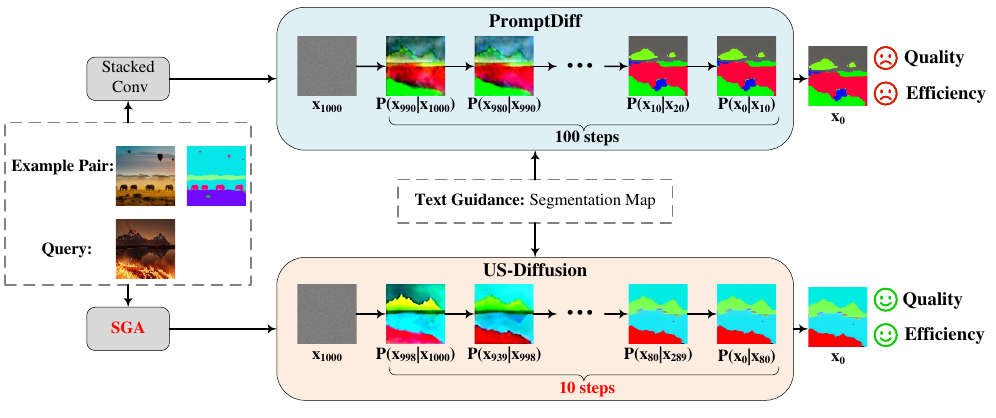}
\caption{Comparison of PromptDiff\cite{promptdiffusion} and our US-Diffusion model. The upper part shows how PromptDiff generates a segmentation map from a query image given an example pair. It requires 100 denoising steps, and the resulting map lacks consistency in structure and object shape with the input query image. The lower part illustrates our US-Diffusion, which generates a segmentation map in just 10 denoising steps, producing a more consistent map in terms of structure and object shape.}
\label{fig_motivation}
\end{figure*}

However, the application of diffusion models in CV presents unique challenges not as prominent in NLP. One of the most notable differences lies in the spatial complexity inherent of visual data\cite{cv5,cv6}. Unlike sequential textual data, visual data requires a deeper understanding of complex spatial structures, diverse modalities, and fine-grained semantics. For example, tasks such as semantic segmentation\cite{semantic1,semantic2} and depth estimation\cite{depth1,depth2} require fine distinctions in shape, size, and spatial relationships, which further compound the difficulty. Furthermore, as the number of tasks and application scenarios in CV continues to grow, existing models\cite{promptdiffusion,instructdiffusion,cv8,instructcv} often struggle with generalization across diverse contexts while maintaining computational efficiency.

A critical challenge faced by current diffusion-based models\cite{promptdiffusion,ipromptdiffusion,controlnet,contextdiffusion} is their limited capacity to effectively capture and leverage the underlying semantic information of visual inputs. These models fail to preserve crucial structural information such as edges, textures, and shapes, which are fundamental to the interpretation of images. As demonstrated in Fig. \ref{fig_motivation}, images generated by PromptDiff\cite{promptdiffusion} frequently exhibit a lack of structural consistency with the input conditions, resulting in outputs that fail to align with expected visual characteristics. This highlights a significant gap in the ability of current models to maintain underlying semantic information awareness during the generation process, which is essential for high-quality results in a wide range of vision tasks.

Another critical limitation of existing diffusion models is their computational inefficiency\cite{stablediffusion}, particularly during the training phase. While diffusion models make significant strides in reducing inference times, their training processes remain computationally expensive. For instance, models such as PromptDiff\cite{promptdiffusion} typically involve a lengthy training process with 1000 noise-adding and denoising steps. During inference, this process is often reduced to mere 100 steps employing  techniques like Denoising Diffusion Implicit Models (DDIM)\cite{ddim}. While this reduction improves inference speed, the disparity between the computational demands of training and inference leads to inefficient utilization of computational resources, which undermines the practicality in practice.

To this end, we introduce Underlying Semantic Diffusion (US-Diffusion), an enhanced framework designed to boost the in-context learning ability of diffusion models. Several novel architectural and methodological improvements significantly boost in-context learning capabilities, understanding of underlying semantic information, and computational efficiency. As shown in Fig. \ref{fig_motivation}, PromptDiff\cite{promptdiffusion} requires 100 time steps to generate an image, while our US-Diffusion is more efficient, needing only 10 time steps to produce an image of better quality. Specifically, we focus on the following key contributions:
\begin{itemize}
	\item We propose Separate \& Gather Adapter (SGA), a tailored adaptation module designed to boost in-context learning for diffusion models. On the one hand, it separately processes the input conditions of different tasks to learn underlying semantic information, enabling the model to better adapt to diverse task-specific contexts. On the other hand, it adopts a shared architecture to improve generalization across tasks, facilitating effective transfer learning and in-context learning on multi-task scenarios.
	\item We develop a novel Feedback-Aided Learning (FAL) framework, which leverages feedback information to provide supervision and guide the model to focus on the underlying semantics of different tasks. This framework boosts the in-context learning by dynamically adjusting its understanding of task-specific contexts during training.
	\item To enhance training efficiency and speed up inference, we propose an Efficient Sampling Strategy (ESS). This strategy involves denser sampling at time steps with higher noise levels during diffusion process, with a total of 10 time steps sampled for both training and inference.
\end{itemize}

\section{Related Work}
\subsection{Image Generation Models}
The field of image generation witnesses remarkable progress through various generative paradigms. Early breakthroughs in Generative Adversarial Networks (GANs)\cite{gan1,gan2,gan3} establish a competitive framework where generator and discriminator networks drive the synthesis of realistic images. While GANs achieve photorealistic quality, their susceptibility to mode collapse and unstable training dynamics motivated the exploration of alternative approaches.

Diffusion models emerge as a transformative solution, leveraging an iterative denoising process to transform random noise into coherent images through a Markov chain. Despite their theoretical advantages, the computational expense of pixel-space diffusion spurs the development of Latent Diffusion Models (LDMs)\cite{stablediffusion}, which operate in a compressed latent space leveraging pre-trained autoencoders\cite{vae2}. Recent advancements aim to enhance LDMs by integrating multimodal control signals, thereby expanding their versatility and applicability. For instance, GLIGEN\cite{gligen} enables open-world grounded text-to-image generation by incorporating spatial inputs such as bounding boxes and keypoints. T2I-Adapter\cite{t2i-adapter} introduces lightweight adapters to align external control signals (e.g., sketches, depth maps) with pre-trained text-to-image models, enabling fine-grained control without modifying the original model weights. InstructCV\cite{instructcv} reformulates computer vision tasks as language-guided image generation problems, utilizing instruction-tuned diffusion models for multi-task generalization. ControlNet\cite{controlnet} enhances pre-trained diffusion models (e.g., Stable Diffusion\cite{stablediffusion}) by introducing spatial conditioning mechanisms, supporting diverse inputs such as edges, poses, and segmentation maps. Similarly, PromptDiff\cite{promptdiffusion}, a vision foundation model, employs diffusion models for in-context learning, enabling flexible adaptation to various visual tasks.

However, these methods remain constrained by the inherent limitations of their LDM foundations. The iterative Gaussian denoising process still demands high computational resources, limiting real-time applicability. Moreover, their reliance on latent-space reconstruction creates an information bottleneck, compromising fine-grained visual details. Our method addresses these challenges through a feedback training framework that introduces direct image-space supervision to complement latent-space optimization, enhancing both generation quality and detail preservation, and an efficient sampling strategy that reduces training and inference steps while maintaining denoising trajectory fidelity, significantly improving computational efficiency without quality degradation. 
\subsection{In-Context Learning in CV}  
In the domain of CV, in-context learning is emerging as a powerful technique to enhance the performance and adaptability of generative models, particularly diffusion models. This approach involves conditioning the generative process on context, such as visual examples, text prompts, and image conditions\cite{promptdiffusion,ipromptdiffusion,contextdiffusion}. By leveraging these contextual elements, models produce more semantically relevant and coherent outputs. This aligns with the growing demand for task-agnostic models capable of generalizing across diverse vision tasks, such as image controllable generation\cite{imagegeneration1,imagegeneration2,stablediffusion,controlnet}, dense predicttion\cite{semantic1,semantic2,depth1,depth2}, image-text retrieval\cite{cv3,cv4}, or visual question answering (VQA)\cite{vqa}, all without requiring specialized modules. For instance, PromptDiff\cite{promptdiffusion} demonstrates the potential of leveraging natural language descriptions and contextual cues to guide the generative process in alignment with user intentions.

However, integrating in-context learning with diffusion models poses challenges, particularly in effectively capturing and utilizing underlying semantics. While additional context improves generation quality, models often struggle to align high-level understanding with fine-grained details required during denoising. For example, complex prompts may lead to outputs lacking precise spatial relationships or visual richness since they fail to accurately reflect intended semantic structures. This highlights the need for more robust mechanisms to enhance the model’s ability to perceive and generate detailed visual content. Therefore, our method introduces Separate \& Gather Adapter (SGA), a tailored adaptation module designed to boost in-context learning in diffusion models by efficiently capturing and leveraging underlying semantics. 
\subsection{Diffusion Model Acceleration}
While diffusion models demonstrate remarkable success in image generation, their reliance on iterative denoising using a U-Net\cite{unet} architecture results in long inference times, which is a significant limitation for time-sensitive applications. To mitigate this, researches focus on accelerating diffusion models through two approaches: training-free and training-based methods.

Training-free methods aim to develop efficient diffusion solvers that reduce sampling steps without additional training. For example, the DDIM\cite{ddim} reformulates the reverse diffusion process of diffusion models as an ordinary differential equation (ODE), reducing sampling steps to 50$\sim$100 without compromising quality, while DPM-Solver\cite{dpm} introduces a high-order solver for diffusion ODEs, further reducing steps to 10$\sim$50. However, these methods fail to sufficiently account for the varying significance of time steps with different noise levels, particularly overlooking the critical role of time steps with high-noise levels in diffusion process. On the other hand, training-based methods often leverage knowledge distillation to accelerate the process. For instance, progressive knowledge distillation\cite{progressive} trains a teacher model with many steps and then distilling it into student models with progressively fewer steps, achieving as few as 10 steps while maintaining quality. Despite their effectiveness, these methods incur higher training costs and may suffer from suboptimal performance with fewer steps. Additionally, they often struggle to generalize across diverse tasks, as distilled models are typically optimized for specific scenarios, limiting their broader applicability.

To address these challenges, we propose a plug-and-play Efficient Sampling Strategy (ESS). It optimizes the sampling process by emphasizing the significance of critical diffusion stages, particularly time steps with high-noise levels, while maintaining high generation quality and computational efficiency. ESS provides a flexible and easily integrable solution, applicable to diverse diffusion-based frameworks, where its effectiveness and universality are demonstrated in \ref{ESS}.

\begin{figure*}[htbp]
\centering
\includegraphics[scale=0.9]{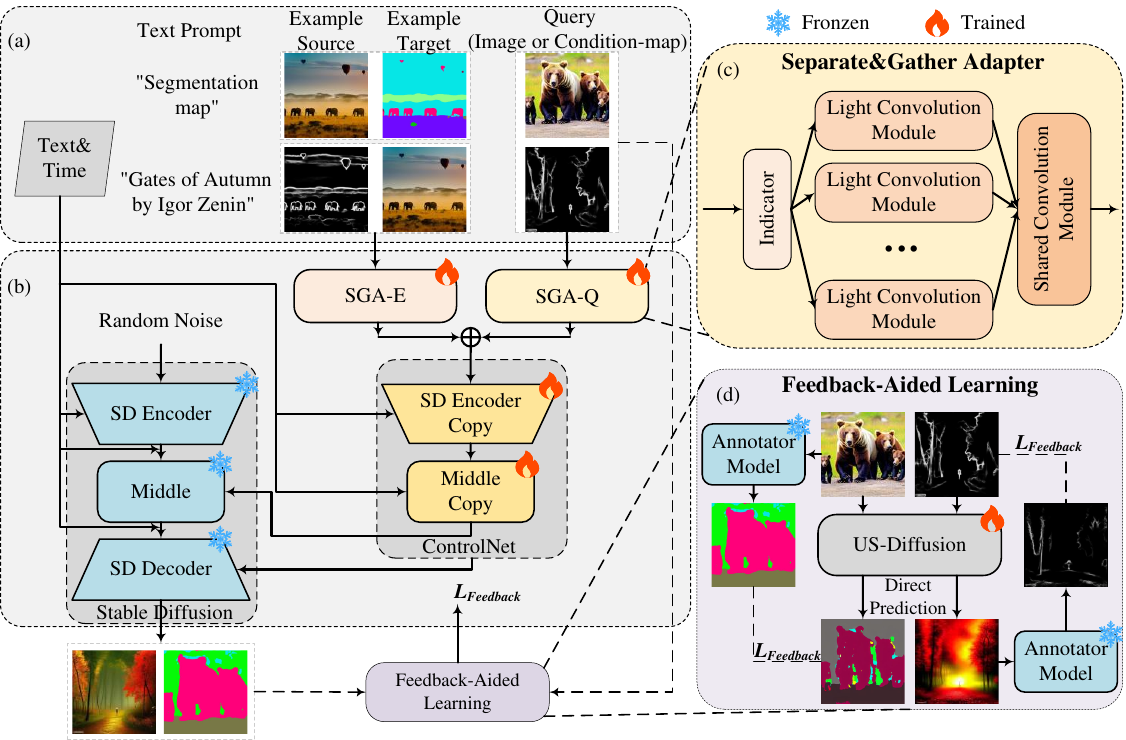}
\caption{The framework of our US-Diffusion model. (a) illustrates the conditional input, including an example pair (example source image and example target image), a query image of the same type as the example source image, and a text prompt. (b) depicts the overall architecture of the model. (c) presents the structure of the proposed Separate \& Gather Adapter (SGA). (d) shows the Feedback-Aided Learning (FAL) framework, which divides the process into two paths to provide task-specific feedback. Notably, a novel Efficient Sampling Strategy (ESS) is proposed during the noise-adding and denoising stages.}
\label{fig_framework}
\end{figure*}
\section{Method}
\subsection{Overall Architecture}
The proposed framework, US-Diffusion, aims to overcome the limitations of current diffusion models in terms of generation quality and computational efficiency. As illustrated in Fig. \ref{fig_framework}, the framework comprises four main components: the widely-used Stable Diffusion (SD) model \cite{stablediffusion}, the ControlNet \cite{controlnet} module equipped with our Separate \& Gather Adapter (SGA), the proposed Feedback-Aided Learning (FAL) framework, and the conditional input component. Together, these components enhance the model's ability to better capture underlying semantic information and improve generation quality. To tackle the issue of slow inference speed that is typical in diffusion models \cite{stablediffusion,controlnet,promptdiffusion}, we also propose an innovative Efficient Sampling Strategy (ESS).

The framework is tailored for two types of tasks: Map2Image and Image2Map, as shown in Fig. \ref{fig_framework}(a). For Map2Image tasks, the model takes a condition-map image (such as a depth map, edge map, or semantic segmentation map) as input, paired with a task-specific example pair and a text prompt like ``Gates of Autumn by Igor Zenin'', to generate a corresponding realistic image. For Image2Map tasks, the model is required to produce the corresponding condition-map image (e.g., a depth map, edge map, or segmentation map) based on the provided realistic image and a task-specific example pair. However, in this task setting, task names such as ``Segmentation map'' are adopted to serve as the text prompts for the Image2Map tasks. Both two types of tasks require the model to effectively capture underlying semantic features, such as edges, textures, and shapes, which are essential for generating high-quality outputs.

Overall workflow begins with an input query (which could be an image or a condition-map, depending on type of task) and an example pair that provides in-context information for task. The query and the example pair are first encoded into latent representations leveraging  a pre-trained encoder\cite{vae2}. Subsequently, these input conditions are processed by our proposed Separate \& Gather Adapter (SGA), and then input into ControlNet, which outputs the control signals embeddings derived from the aforementioned condition information to the reverse process of Stable Diffusion\cite{stablediffusion}. The final output is obtained by iteratively denoising the latent representation via the reverse process, guided by embeddings from ControlNet and text embeddings.  

The following sections will provide a detailed explanation of each component, outlining how they are implemented and their contributions to our overall framework. Section \ref{Preliminary} introduces preliminary knowledge, Section \ref{SGA} explains the SGA, Section \ref{FAL} covers the introduction of FAL, and Section \ref{ESS} describes the novel ESS approach.

\subsection{Preliminary}
\label{Preliminary}
Denoising Diffusion Probabilistic Model (DDPM)\cite{ddpm} is a deep generative model that involves two Markov chains, referred to as the forward diffusion process and the reverse process. Each Markov chain spans $T$ steps, called diffusion steps or reverse steps.

Forward diffusion process gradually introduces Gaussian noise into the data distribution until the noisy data distribution approximates a Gaussian distribution. Formally, the diffusion process from $x_{0}\sim q_{d ata}$ to $x_{T}$ is defined as:
\begin{equation}
q(x_{ 1 },\cdots,x _{T}|x_{0}) = \prod_{i=1}^{T}q(x_{t}|x_{t-1}),
\end{equation}
where each $q(x_{t}|x_{t-1})$ is defined as $N(x_{t};\sqrt{1-\beta_{t}}x_{t-1}, \break\beta_{t}\textbf{I})$. The hyperparameters $\beta_{1}, \beta_{2}, \cdots, \beta_{T} \ (\beta_{t}>0)$  are referred to as the variance schedule.

Reverse process aims to remove the noise added at each step of the forward diffusion. Formally, the reverse process from $x_{T}$ to $x_{0}$ is defined as:
\begin{equation}
p _ { \theta } ( x _ { 0 } , \cdots , x _ { T - 1 } | x _ { T } ) = \prod _ { t = 1 } ^ { T } p _ { \theta } ( x _ { t - 1 } | x _ { t } ),
\end{equation}
where each $ p _ { \theta } ( x _ { t - 1 } | x _ { t } )$ is defined as $ N ( x _ { t - 1 } ; \mu _ { \theta } ( x _ { t } , t ) , \break\sigma _ { t } ^ { 2 } \textbf{I} )$. The mean $\mu _ { \theta } ( x _ { t } , t )$ is parameterized by a neural network, and the variance $\sigma _ { t }$ is a constant dependent on the time step. Based on the reverse process, the sampling procedure starts by sampling a random Gaussian noise $ x _ { T } \sim N ( 0 , \textbf{I} )$ , followed by sampling $ x _ { t - 1 } \sim p _ { \theta } ( x _ { t - 1 } | x _ { t } )$ for $ t = T , T - 1 , \cdots , 1 ,$ and finally outputs $x_{0}$.

The training objective of DDPM, derived from the Evidence Lower Bound (ELBO), simplifies significantly under the parameterization introduced by \cite{ddpm}, enabling direct sampling of \( x_{t} \) from \( x_{0} \) as follows:
\begin{equation}
  q ( x _ { t } | x _ { 0 } ) = N ( x _ { t } ; \sqrt { \overline { \alpha } } _ { t } x _ { t - 1 } , ( 1 - \overline { \alpha } _ { t } ) \textbf{I} ),
\end{equation}
where \( \overline{\alpha}_{t} = \prod_{i=1}^{t} \alpha_{i} \), and \( \alpha_{t} = 1 - \beta_{t} \), representing the noise scheduler \( \alpha \).

Additionally, the mean $\mu _ { \theta } ( x _ { t } , t ) $ is parameterized as:
\begin{equation}
\mu _ { \theta } ( x _ { t } , t ) = \frac { 1 } { \sqrt { \alpha _ { t } } } ( x _ { t } - \frac { \beta _ { t } } { \sqrt { 1 - \alpha _ { t } } } \epsilon_ { \theta } ( x _ { t } , t ) ),
\end{equation}
where $\epsilon_ { \theta }$ is a noise-prediction model taking $x_{t}$ and the diffusion step $t$ as inputs. Furthermore, $\sigma_{t}$ is simply parameterized as $ \tilde { \beta } _ { t } ^ { \frac { 1 } { 2 } }$, where $ \tilde { \beta } _ { t } = \frac { 1 - \overline { \alpha } _ { t - 1 } } { 1 - \overline { \alpha } _ { t } } \beta _ { t }$, $ \tilde { \beta } _ { 1 } = \beta _ { 1 }$. The training objective of DDPM is to minimize the following loss function:
\begin{equation}
  L_{\text{train}} = 
  \mathbb{E}_{\substack{x_{0} \sim p_{\text{data}} \\ \epsilon \in N(0,\textbf{I}) \\ i \in [1, \cdots, T]}}
  \left[ \| \epsilon - \epsilon_{\theta}(x_{t},t) \|_{2}^{2} \right],
\end{equation}
where $T$ is the total number of time steps, $t =i/T$, and $ x _ { t } = \sqrt { \overline { \alpha } } _ { t } \cdot x _ { 0 } + \sqrt { 1 - \overline { \alpha } _ { t } } \cdot \epsilon$, derived from the above property. This objective is interpreted as minimizing the mean squared error between the true noise $\epsilon$ and the predicted noise $ \epsilon _ { \theta } ( x _ { t } , t )$ at each time step.

Initially, this approach is implemented in pixel-space, but now it is extended to the latent space for faster inference and improved quality. Here prior to the diffusion process, image $x$ is encoded by an encoder $ \varepsilon$ into the latent image $z _{x}$, which is subsequently decoded by a decoder $D$ to convert the latent image to the image space. However, the model $\epsilon_{\theta}(x_{t},t)$ is unconditional, which means the reverse process is not controllable. To guide this process, an additional condition $c$ ($c$ is the conditional embedding that is processed from text or a task-specific modality) is provided to the noise-prediction model $\epsilon_{\theta}$, leading to the following modified training loss:

\begin{equation}
  L_{\text{train}} = 
  \mathbb{E}_{\substack{x_{0} \sim p_{\text{data}} \\ \epsilon \in N(0,\textbf{I}) \\ i \in [1, \cdots, T]}}
  \left[ \| \epsilon - \epsilon_{\theta}(z_{x_{t}}, t, c_{t}, c_{f}) \|_{2}^{2} \right],
\end{equation}
where $ z _ { x _ { t } }=\varepsilon ( x _ { t } )$,  $ z _ { x_{t} } = \sqrt { \overline { \alpha } } _ { t } \cdot z _ { x_{0} } + \sqrt { 1 - \overline { \alpha } _ { t } } \cdot \epsilon$, $c_{t}$ is a text embedding and $c_{f}$ is a task-specific embedding that is spatially aligned with an image in general. This objective is interpreted as minimizing the mean squared error between the true noise $\epsilon$ and the predicted noise $\epsilon_{\theta}(z_{x_{t}}, t, c_{t}, c_{f})$ at each time step.

\subsection{Separate \texorpdfstring{$\&$}{\&} Gather Adapter}
\label{SGA}
Existing diffusion-based approaches, such as PromptDiff\cite{promptdiffusion}, advance multi-task image generation, including depth estimation, edge detection, semantic segmentation and image controllable generation. However, a key limitation is their inability to effectively capture and leverage underlying semantic information, such as edges, textures, and shapes. These features are crucial for generating high-quality, semantically meaningful images. Without robust mechanisms to extract and utilize such information, generated images may lack fine-grained details or fail to accurately reflect intended visual structures, especially under diverse task-specific conditions.

To address this, we propose a novel Separate \& Gather Adapter (SGA), illustrated in Fig. \ref{fig_framework}(c). SGA boosts the model's in-context learning by adapting to underlying semantic visual information across varying task conditions. At input stage, an indicator function separates tasks and assigns each to a dedicated lightweight convolutional module. These task-specific modules, composed of stacked lightweight convolutional layers, efficiently capture task-relevant features while maintaining computational efficiency. By isolating task-specific processing, the adapter ensures the model focuses on the unique characteristics of each task, boosting in-context learning in diffusion models.

At output stage, SGA gathers outputs from all task-specific modules and processes them through a shared convolutional module. This shared module acts as a feature integrator, combining diverse features into a unified representation. It captures common underlying semantic information, such as spatial relationships, structural patterns, and fine-grained details, across tasks. By integrating task-specific and shared features, SGA enables the model to generate images tailored to specific task requirements while preserving rich semantic content, thereby enhancing overall image quality. Formally, the operation of SGA is expressed as:
\begin{equation}
F _ {SGA} = \sum _ { i = 1 } ^ { K } \mathbb{I} ( i = k ) F _ { s h a r e d - c o n v } ( F _ { l i g h t - c o n v } ^ { ( i ) } ( I^k) ),
\end{equation}
where \( I^k \) represents the input images or condition-map images associated with task \( k \). \( F _ { l i g h t - c o n v } ^ { ( i ) } \) denotes the \( i \)-th lightweight convolutional module, which processes the input conditions specific to task \( i \). \( F _ { s h a r e d - c o n v } \) is a shared convolutional module that gathers the outputs from all task-specific modules, ensuring that the model captures both task-specific and shared semantic features. $\mathbb{I}(\cdot)$ is a indicator function that separates input tasks and assigns each task to its corresponding lightweight convolutional module. The indicator function $\mathbb{I}(\cdot)$ is defined as:
\begin{equation}
\mathbb{I}(i = k) = 
\begin{cases} 
1, & \text{if } i = k \\
0, & \text{if } i \neq k.
\end{cases}
\end{equation}

In practice, as shown in Fig. \ref{fig_framework}(b), we deploy two SGA variants, SGA-E and SGA-Q, to handle distinct information streams. SGA-E processes concatenated pairs of example source and target images (${I_{\text{src}}, I_{\text{tgt}}}$), extracting relationships that encode in-context task priors. Simultaneously, SGA-Q directly processes query images ($I_{\text{query}}$), focusing on their underlying semantic structure. This design allows explicit modeling of both example-guided task priors and query-specific features. The outputs of both variants are summed and input into ControlNet, as follows:
\begin{equation}
F_{\text{ControlNet-Input}} = \underbrace{F_{\text{SGA-E}}(I^k_{\text{src}} \oplus I^k_{\text{tgt}})}_{\text{Example guidance}} + \underbrace{F_{\text{SGA-Q}}(I^k_{\text{query}})}_{\text{Query features}},
\end{equation}
where $\oplus$ denotes channel-wise concatenation. Each SGA variant maintains distinct processing streams:
\begin{equation}
F_{\text{SGA-E}} = \sum_{i=1}^{K} \mathbb{I}(i=k) F_{\text{shared-conv}}^{(\text{E})}\left(F_{\text{light-conv}}^{(i,\text{E})}(I^k_{\text{src}} \oplus I^k_{\text{tgt}})\right),
\end{equation}
\begin{equation}
F_{\text{SGA-Q}} = \sum_{i=1}^{K} \mathbb{I}(i=k) F_{\text{shared-conv}}^{(\text{Q})}\left(F_{\text{light-conv}}^{(i,\text{Q})}(I^k_{\text{query}})\right).
\end{equation}

In conclusion, the SGA boosts in-context learning by introducing SGA-E, which extracts example-based task priors, and SGA-Q, which focuses on query-specific features. By summing their outputs and integrating them into ControlNet, our model combines example guidance and query-specific features, enabling high-quality image generation that is both visually coherent and semantically rich. This approach improves task adaptability and maintains computational efficiency through lightweight convolutional modules and shared architecture components. The SGA represents a significant advancement in multi-task image generation, capturing fine-grained details and semantic structures while boosting in-context learning in diffusion models.
\subsection{Feedback-Aided Learning}
\label{FAL}
Current diffusion models face challenges in generating high-quality visuals, particularly with object structure distortions. These issues arise from relying solely on reconstruction loss in the latent space, without supervision based on visual perception in the image space. To address this, we propose the Feedback-Aided Learning (FAL) framework, as shown in Fig. \ref{fig_framework}(d). FAL boosts in-context learning by providing feedback supervision that guides the model to focus on underlying semantic information of different tasks, improving its ability to generalize in specific contexts.

The FAL framework leverages annotation models employed in ControlNet\cite{controlnet}, which generates condition-map images from input images. Feedback between the condition-map image generated by our model (or extracted from  output image) with the corresponding condition-map image extracted from input image (or the provided condition-map input), helping our model focus on task-relevant semantic information. This feedback loop boosts the model’s in-context learning by refining its understanding of the task through iterative adjustments.

To further enhance efficiency of the feedback learning, we depart from the traditional sequential refinement of predictions from \(x_{T}\) to \(x_{0}\) and instead make direct predictions at intermediate step \(t \in \left[1, t^{\prime}\right]\) (with \(t^{\prime} = 200\) in our work). Additionally, depending on the task—Image2Map or Map2Image—the FAL framework integrates two distinct feedback paths, as shown in Fig. \ref{fig_framework}(d).

\subsubsection{\textbf{Feedback-Aided Learning for Image2Map Tasks}} For Image2Map tasks where input query is a image and output is a condition-map image, we first adopts the annotation model employed in ControlNet\cite{controlnet} to extract the corresponding condition-map image from the query image. At output, the innovate direct prediction approach is applied to obtain the output condition-map image. The feedback loss $L_{Feedback}$ is then computed as the MSE loss between the output condition-map image and the extracted condition-map image in image space rather than latent space, guiding the model's fine-tuning. The $L_{Feedback}$ for this task setting is as follows:
\begin{equation}
L _ { F e e d b a c k } = L ( D ( x _ { q u e r y } ) , G _ { t ^ { \prime } } ( x _ { t } , t , c _ { t } , c _ { f } ) ),
\end{equation}
where $x_{query}$ is the input query image, $D$ is the annotation model employed in ControlNet to extract the condition-map images from images. $G _ { t ^ { \prime } }$ represents the innovate direct prediction approach is applied to obtain the output, which generates the direct prediction outcome $x _ { t } \rightarrow x _ { 0 } ^ { \prime }$ at  intermediate step $t \in \left[ 1 , t ^ { \prime } \right]$ based on the condition information embedding $c_{f}$ from ControlNet and the text guidance embedding $c_{t}$.
\subsubsection{\textbf{Feedback-Aided Learning for Map2Image Tasks}} For Map2Image tasks where input query is a condition-map image and output is an image, we apply the innovate direct prediction approach at output to obtain the image. Then, we adopts the annotation model employed in ControlNet to extract the corresponding condition-map image from the output image. Since the input query image is already a condition-map image, no additional processing is required. Finally, $L_{Feedback}$ is computed as the MSE loss between the input condition-map image and the extracted condition-map image in image space rather than latent space, guiding the model's fine-tuning. The $L_{Feedback}$ for this task setting is as follows:
\begin{equation}
L _ { F e e d b a c k } = L ( x _ { q u e r y } , D ( G _ { t ^ { \prime } } ( x _ { t } , t , c _ { t } , c _ { f } ) ) ),
\end{equation}
where $x_{query}$ is the input condition-map image, $D$ is the annotation model employed in ControlNet to extract the condition-map images from images. $G _ { t ^ { \prime } }$ represents the innovate direct prediction method applied to obtain the output, which generates the direct prediction outcome $x _ { t } \rightarrow x _ { 0 } ^ { \prime }$ at  intermediate step $t \in \left[ 1 , t ^ { \prime } \right]$ based on the condition information embedding $c_{f}$ from ControlNet and the text guidance embedding $c_{t}$.

Then, we insert the $L_{Feedback}$ into our training loss. The total loss is expressed as:
\begin{equation}
L _ { T o t a l } = L _ { T r a i n } + \lambda L _ { F e e d b a c k },
\end{equation}
where $\lambda$ represents the weight of the $L_{Feedback}$.

By integrating feedback supervision in image space into the training process of our model, FAL not only improves the model's ability to generate high-quality outputs but also boosts its in-context learning capabilities, enabling it to better adapt to and generalize within specific task contexts. This approach ensures that the model remains focused on the underlying semantic information, leading to more accurate and contextually relevant predictions.
\subsection{Efficient Sampling Strategy}
\label{ESS}
While addressing generation quality through previous components, the inherent low inference
speed of diffusion models remains a critical challenge. To mitigate this inefficiency, we propose an Efficient Sampling Strategy (ESS) that innovatively optimizes both the time-step sampling mechanism and training process. The core insight behind our approach stems from two key observations: 1) High-noise stages in the diffusion process contribute significantly to generation quality, as these phases are crucial for shaping the overall structure and details of the generated images. However, conventional uniform sampling does not prioritize these critical phases, leading to inefficient allocation of computational resources, and 2) Existing acceleration methods often introduce a mismatch between the time steps adopted during training and inference, resulting in suboptimal utilization of learned parameters.

To address these issues, we introduce ESS that enhances the time-step sampling mechanism employed by the model's noise scheduler.
\begin{figure}[!t]
  \centering
  \includegraphics[scale=0.4]{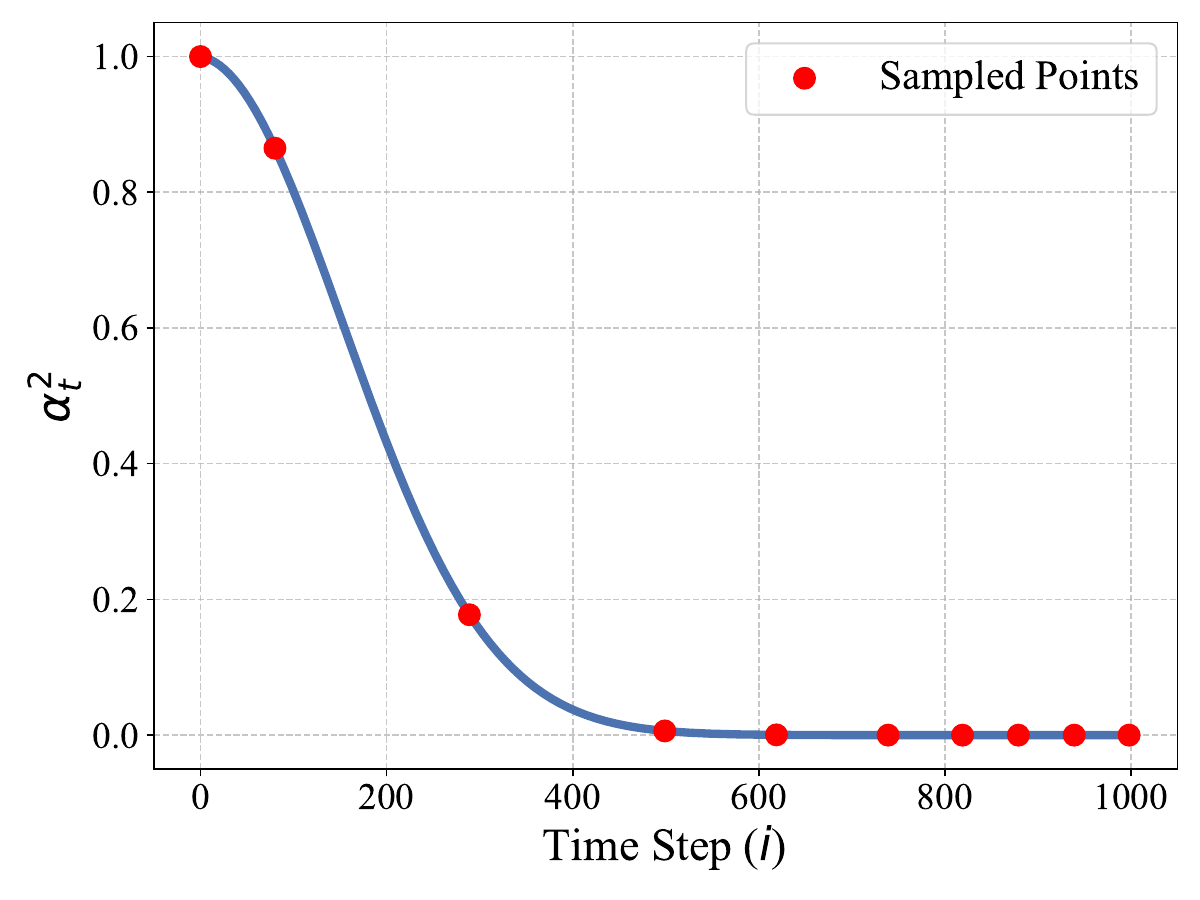}
  \caption{The curve of $\alpha^ { 2 } ( t )$ ($t=i/1000$), where 10 discrete time steps out of the 1000 discrete time steps are non-uniformly sampled.}
  \label{fig_non}
\end{figure}
The ESS we proposed involves non-uniform sampling of certain time steps within $\left[0,T\right]$ during the inference phase, with denser sampling at time steps that are at a higher noise level. Additionally, since the inference phase is conducted only at these time steps, to enhance training efficiency, we assume that it is not necessary to compute the mean square error loss between the true noise $\epsilon$ and the predicted noise $\epsilon_{\theta}(z_{x_{t}},t,c_{t},c_{f})$ at every time step, but rather only at these sampled time steps. 

In DDPM\cite{ddpm}, the corresponding $\alpha$ noise scheduler $\alpha(i/T)$ $(T=1000$) at the discrete time step $i$ is give by:
\begin{equation}
\alpha ( \frac { i } { T } ) = \prod _ { j = 1 } ^ { i } \left( 1 - \beta ( \frac { j } { T } ) \right),
\end{equation}
where $i/T$ is denoted as $t$, and then we obtain the continuous function form of the $\alpha$ noise scheduler $\alpha(t)$.

In this paper, in order to visually demonstrate our efficient sampling strategy, we first plot a curve of $\alpha^ { 2 } ( t )$ and then we sample 10 discrete time steps out of the 1000 discrete time steps non-uniformly, as shown in the Fig. \ref{fig_non}.

\section{Experiments}
\subsection{Datasets and Evaluation Metrics}
We employ the CLIP-filtered Instruct-Pix2Pix dataset\cite{instructpix2pix} to train our model followed by PromptDiff\cite{promptdiffusion}. Specifically, this dataset is a Stable Diffusion\cite{stablediffusion} synthetic image editing prompt dataset, consisting of 313,010 image editing pairs. Additionally, all the condition-map images (e.g., depth maps) utilized in this work are generated via the annotation models employed in ControlNet\cite{controlnet}. Following PromptDiff, we allocate five percent of the total data as a test set for evaluation, two types of training tasks include: Map2Image tasks (Seg-to-Image, Hed-to-Image, Depth-to-Image) and Image2Map tasks (Image-to-Seg, Image-to-Hed, Image-to-Depth). Additional evaluation tasks in this dataset are employed for verifying the generalization of our method, including Canny-to-Image, Scribble-to-Image, Image-to-Canny, and Image-to-Scribble.
For Map2Image tasks, the goal is to reconstruct realistic images from condition-map inputs, such as segmentation maps, edge maps, or depth maps. As shown in Fig. \ref{fig_framework}, this process involves utilizing a task-specific example pair and a text prompt, such as ``Gates of Autumn by Igor Zenin.'' In contrast, Image2Map tasks focus on generating condition-map images (e.g., segmentation maps, edge maps, or depth maps) from realistic images. In these tasks, the text prompts are typically task-specific, such as ``Segmentation map''. 

In order to assess the generalization capability of our model across different tasks and datasets, we expand the evaluation on the MultiGen-20M\cite{unicontrol}, NYUv2\cite{nyu}, and ADE20K\cite{ade1, painter} datasets. 
The MultiGen-20M dataset contains approximately 20 million pairs of high-quality real image-map pairs.
The NYUv2 dataset comprises 464 indoor scenes captured by Microsoft Kinect cameras, providing rich depth information suitable for evaluating image-to-depth tasks. 
The ADE20K dataset includes 25,000 images spanning 150 semantic categories, making it highly suitable for evaluating image-to-seg tasks. 
We randomly select 5000 images along with their corresponding captions from MultiGen-20M, resulting in a test set of 5000 images for evaluation. 
Additionally, we evaluate the performance of our model on the validation sets of the NYUv2 and ADE20K datasets. 

To assess the quality of generation, we follow the evaluation protocol adopted in PromptDiff\cite{promptdiffusion}. For the Map2Image tasks, we provide quantitative results leveraging the reconstructed Fréchet Inception Distance (FID), which measures the distribution similarity between generated and real images using features from a pre-trained Inception-v3 network\cite{inception}. For the Image2Map tasks, we provide quantitative results measured by the Root Mean Square Error (RMSE), a pixel-level geometric accuracy metric that calculates the square root of the average squared differences between predicted and ground truth maps. Specifically, we generate 5,000 images in total to compute these evaluation metrics. Please note that a smaller score for these two metrics indicates better performance.
\subsection{Implementation Details}
Our US-Diffusion framework builds upon the official implementations of PromptDiff\cite{promptdiffusion}. The framework retains the original text encoder and diffusion U-Net from Stable Diffusion v1.5\cite{stablediffusion}, freezing their parameters during training. Only the ControlNet\cite{controlnet} parameters and the proposed SGA are updated. We employ the AdamW\cite{adamw} optimizer with a fixed learning rate of \(1 \times 10^{-4}\) for optimization. The batch size is set to 8, allowing the model to process a sufficient number of samples per iteration while maintaining manageable memory usage. Training is firstly conducted for 10,000 steps on a single NVIDIA 4090 GPU, ensuring that the model has ample opportunity to learn the intricate patterns present in the dataset. To further investigate the impact of batch size on model performance, we also conduct experiments with an increased batch size of 16 on a single A800 GPU. This adjustment aims to explore whether increasing the batch size will lead to improved performance and potentially enhance the model's generalization capabilities.
Please note that our model only be trained on the six training tasks with Instruct-Pix2Pix dataset, and other datasets are only employed for evaluation.
\begin{table*}[htbp]
\renewcommand{\arraystretch}{1.2}
\caption{Comparisons of Different Methods on \textbf{Instruct-Pix2Pix Dataset}. US-Diffusion* (Ours) Refers to the Model Is Trained With the Batch Size of 16\label{table_indata}}
\centering
\begin{tabular}{c c c c p{0.01cm} c c c}
\toprule
\multirow{3}{*}{Model} & \multicolumn{3}{c}{Map2Image (FID $\downarrow$)} & & \multicolumn{3}{c}{Image2Map (RMSE $\downarrow$)} \\  
\cmidrule(r){2-4}
\cmidrule(r){6-8}
 & Seg-to-Image & Hed-to-Image & Depth-to-Image && Image-to-Seg & Image-to-Hed & Image-to-Depth \\  
\midrule                 
GLIGEN\cite{gligen} &27.48 &34.10 &18.22&&-&-&- \\
T2I-Adapter\cite{t2i-adapter} &28.13 &- &18.59&&-&-&- \\
Unified-IO\cite{unified-io} & -&- &-&& 0.375&-&0.353 \\
InstructCV\cite{instructcv} &- &- &-&&0.499&-&0.422 \\
ControlNet (FT)\cite{controlnet} & 29.58 & 21.36 & 25.46 && 0.380 & 0.240 & 0.298 \\
PromptDiff\cite{promptdiffusion} & 25.32 & 23.72 & 25.26 && 0.406 & \textbf{0.196} & 0.272 \\
US-Diffusion (Ours) & \underline{18.21} & \underline{16.62} & \underline{17.05} && \underline{0.350} & 0.207 & \underline{0.239} \\
US-Diffusion* (Ours) & \textbf{16.17} & \textbf{16.08} & \textbf{15.91} && \textbf{0.348} & \underline{0.197} & \textbf{0.221} \\
\bottomrule          
\end{tabular}
\end{table*}

\subsection{Comparison with State-of-the-Art Methods}
In this section, we evaluate our model's performance against state-of-the-art methods across various tasks. 
First, we compare our model on the six training tasks, as shown in Table \ref{table_indata}. For fairness, we replicate PromptDiff's results executing their open-sourced code under identical conditions. Then, we extend our evaluation by conducting comparisons for the Map2Image and Image2Map tasks on a broader dataset, MultiGen-20M\cite{unicontrol}, while also performing extra assessments on the NYUv2 \cite{nyu} and ADE20K \cite{ade1,painter} datasets for the Image2Map tasks, as shown in Table \ref{table_compare1}-Table \ref{table_compare2}. Finally, we assess our model's generalization on novel tasks through the Instruct-Pix2Pix dataset, providing insights into its adaptability and robustness on unseen tasks, as shown in Table \ref{table_new_tasks}.

All the comparison methods in this section including GLIGEN\cite{gligen}, T2I-Adapter\cite{t2i-adapter}, Unified-IO \cite{unified-io}, InstructCV \cite{instructcv}, ControlNet \cite{controlnet}, and PrompDiff\cite{promptdiffusion}. 
GLIGEN\cite{gligen} is an open-world text-to-image generation method that incorporates spatial inputs such as bounding boxes and keypoints, supporting the Map2Image task type.
T2I-Adapter\cite{t2i-adapter} introduces lightweight adapters to align external control signals (e.g., sketches, depth maps) with pre-trained text-to-image models, enabling fine-grained control without modifying the original model weights, supporting the Map2Image task type.
Unified-IO\cite{unified-io} adopts a sequence-to-sequence framework, unifying diverse inputs and outputs into discrete tokens to handle a wide range of AI tasks across vision and language domains, supporting the Image2Map task type.
InstructCV\cite{instructcv} transforms computer vision tasks into language-guided image generation problems, leveraging instruction-tuned diffusion models for multi-task generalization, supporting the Image2Map task type.
ControlNet \cite{controlnet} enhances pre-trained image diffusion models with task-specific conditions (e.g., depth maps), supporting various inputs such as edges, depth, and segmentation maps. The original ControlNet supports only the Map2Image task type.
In this work, we fine-tune ControlNet separately on our six training tasks from the Stable Diffusion \cite{stablediffusion} checkpoint on the Instruct-Pix2Pix \cite{instructpix2pix} dataset, naming this variant ControlNet (FT), which supports both Map2Image and Image2Map task types.
PromptDiff\cite{promptdiffusion} is a vision foundation model that leverages diffusion models for in-context learning, enabling flexible adaptation to various visual tasks, supporting both Map2Image and Image2Map task types. 

Please note that, except for ControlNet and PromptDiff, which are re-trained by us with the same batch size, all other comparison methods are tested utilizing their publicly available model weights. 

Based on the results from Tables \ref{table_indata}-\ref{table_new_tasks}, we observe the following observations and conclusions:
\subsubsection{\textbf{The Results for Map2Image Tasks}}
In the field of image controllable generation, the Map2Image task plays a pivotal role in assessing a model's capability to reconstruct realistic images from condition-map images, which incorporate elements such as lines, shapes, shadows/light, and depth. 

As shown in Table \ref{table_indata}, our model achieves significantly better results than other methods with a training batch size of 8. Specifically, it outperforms ControlNet (FT)\cite{controlnet} by large margins, with FID score reductions of 11.37 on Seg-to-Image, 4.74 on Hed-to-Image, and 8.41 on Depth-to-Image tasks. Compared with the existing best method, PromptDiff\cite{promptdiffusion}, our model also shows substantial gains, with FID score reductions of 7.11 on Seg-to-Image, 7.10 on Hed-to-Image, and 8.21 on Depth-to-Image tasks. Furthermore, our model outperforms GLIGEN\cite{gligen} and T2I-Adapter\cite{t2i-adapter}, which are specifically designed for Map2Image tasks. These improvements are driven by enhanced in-context learning enabled by our proposed SGA and FAL. SGA enhances the model's ability to capture fine-grained semantic structures and spatial relationships in condition maps, while FAL iteratively refines outputs through intermediate feedback, particularly beneficial for tasks requiring precise spatial relationships.

Additionally, increasing training batch size from 8 to 16 further improves performance,  with a maximum FID score reduction of 2.04 for Seg-to-Image task and a minimum FID score reduction of 0.54 for Depth-to-Image task. This is due to more stable gradient estimates, which enhance SGA's ability to integrate shared semantic information and improve FAL's refinement with more diverse examples per iteration. These results highlight the synergy between SGA and FAL, validating their effectiveness in multi-task controllable generation.

\begin{table*}[htbp]
\renewcommand{\arraystretch}{1.2}
\caption{Comparisons of Different Methods on \textbf{MultiGen-20M}. Symbol * Refers to the Model Is Trained With the Batch Size of 16 \label{table_compare1}}
\centering
\begin{tabular}{c c c c p{0.01cm} c c c}
\toprule
\multirow{3}{*}{Model} & \multicolumn{3}{c}{Map2Image (FID $\downarrow$)} & & \multicolumn{3}{c}{Image2Map (RMSE $\downarrow$)} \\  
\cmidrule(r){2-4}
\cmidrule(r){6-8}
 & Seg-to-Image & Hed-to-Image & Depth-to-Image && Image-to-Seg & Image-to-Hed & Image-to-Depth \\     
\midrule                
GLIGEN\cite{gligen} &30.80 &41.77 &\underline{18.75}&&-&-&-\\
T2I-Adapter\cite{t2i-adapter} &\underline{27.41} &- &\textbf{17.96}&&-&-&- \\
Unified-IO\cite{unified-io} &-&-&-&&0.501&-&0.332 \\
InstructCV\cite{instructcv} &-&-&-&&0.535&-&0.437 \\
ControlNet* (FT)\cite{controlnet} &33.90&29.28&35.42&&\underline{0.377}& 0.252&0.299 \\
PromptDiff* \cite{promptdiffusion} &36.08 &\underline{28.15} &33.20&&0.422 &\underline{0.232} &\underline{0.292}\\
US-Diffusion* (Ours)  &\textbf{25.49} &\textbf{23.09} &24.15&&\textbf{0.360} &\textbf{0.218} &\textbf{0.229} \\
\bottomrule           
\end{tabular}
\end{table*}
\subsubsection{\textbf{The Results for Image2Map Tasks}}
Compared with the Map2Image tasks, the Image2Map tasks are generally more challenging, primarily because Stable Diffusion\cite{stablediffusion} and other vision models are rarely pre-trained on such specialized data. Another difficulty lies in the text encoder of the original Stable Diffusion. While employing image captions as text prompts is straightforward in the Map2Image tasks, handling prompts for the Image2Map tasks requires careful design. Full image captions may be irrelevant to the specific task, and overly explicit task descriptions risk overfitting, potentially impairing the model's ability to generalize to unseen tasks. Following the evaluation protocol of PromptDiff\cite{promptdiffusion}, we continue to adopt task names as the text prompts for the Image2Map tasks. 

As shown in Table \ref{table_indata}, with a training batch size of 8, our model outperforms other methods on the Image-to-Seg and Image-to-Depth tasks. For Image-to-Seg task, it achieves RMSE score reductions of 0.030 compared with ControlNet (FT)\cite{controlnet} and 0.056 compared with PromptDiff. For Image-to-Depth task, it reduces RMSE score by 0.059 over ControlNet (FT) and 0.033 over PromptDiff. These improvements are driven by the enhanced in-context learning capabilities of SGA and FAL. SGA improves the model's understanding of semantic relationships between the query image and the target condition map, while FAL iteratively refines outputs through intermediate feedback, ensuring higher precision in tasks requiring fine-grained spatial understanding.

For Image-to-Hed task, our model achieves a 0.033 lower RMSE score compared to ControlNet (FT) but a 0.011 higher RMSE score than PromptDiff. However, increasing the training batch size from 8 to 16 further reduces the RMSE score by 0.010 on this task. This improvement is likely due to the nature of Image-to-Hed task, which focuses on extracting object boundaries while excluding textures. A smaller batch size may cause the model to overemphasize local details, limiting its ability to capture global context. A larger batch size enables the model to learn from a broader range of features, enhancing performance on tasks requiring global understanding. This also demonstrates the importance of batch size for tasks requiring global context.
\begin{table}[htbp]
\renewcommand{\arraystretch}{1.2}
\caption{Comparison of Our method with Unified-IO and InstructCV on the \textbf{NYUv2} and \textbf{ADE20K}
\label{table_compare2}}
\centering
\begin{tabular}{c c c}
\toprule
\multirow{3}{*}{Model} &\multicolumn{2}{c}{RMSE $\downarrow$}\\ 
\cmidrule(r){2-3}
 &NYUv2 & ADE20K\\ 
&Image-to-Depth & Image-to-Seg \\      
\midrule                 
Unified-IO\cite{unified-io} &0.385 &0.500 \\
InstructCV\cite{instructcv} &\underline{0.297} &0.491  \\ 
PromptDiff\cite{promptdiffusion} &0.313 & \underline{0.414} \\
US-Diffusion (Ours) &\textbf{0.289} & \textbf{0.370} \\
\bottomrule             
\end{tabular}
\end{table}
\subsubsection{\textbf{Comparison with Other Methods on New Datasets}}
To further validate the generalization ability of our model, we conduct comparisons on the MultiGen-20M\cite{unicontrol}, as shown in Table \ref{table_compare1}. For Image2Map tasks, our model surpasses all other methods. Specifically, it achieves RMSE scores reductions of 0.017, 0.034, and 0.070 on Image-to-Seg, Image-to-Hed, and Image-to-Depth tasks, respectively, compared with ControlNet (FT)\cite{controlnet}. Against the existing best method, PromptDiff\cite{promptdiffusion}, our model reduces RMSE score by 0.062, 0.014, and 0.063 on the same tasks. Additionally, our model significantly outperforms Unified-IO\cite{unified-io} and InstructCV\cite{instructcv}, which are both limited to supporting primarily the Image2Map task type.

For Map2Image tasks, our model also excels, except on the Depth-to-Image task. It achieves FID score reductions of 8.41 on Seg-to-Image task and 6.19 on Hed-to-Image task compared to ControlNet (FT). Against PromptDiff, our model reduces FID score by 10.59 on Seg-to-Image task and 5.06 on Hed-to-Image task. For the Depth-to-Image task, while GLIGEN\cite{gligen} and T2I-Adapter\cite{t2i-adapter} perform slightly better due to their specialized architectures, our model remains competitive without task-specific customization. This highlights the strength of our general-purpose framework, which maintains robust performance across diverse tasks rather than optimizing for a single domain. The balanced performance underscore its versatility and adaptability, making it a comprehensive solution for multi-task generation challenges.

Next, we compare our model with PromptDiff, Unified-IO, and InstructCV on the validation datasets of ADE20K\cite{ade1,painter} and NYUv2\cite{nyu}. Since InstructCV and Unified-IO do not support the Image-to-Hed task, the comparison is limited to Image-to-Seg and Image-to-Depth tasks. Notably, our model is not trained on ADE20K or NYUv2, whereas Unified-IO and InstructCV are. From Table \ref{table_compare2}, we observe that despite not being trained on these datasets, our model outperforms both Unified-IO and InstructCV. Specifically, on the NYUv2 dataset for the Image-to-Depth task, our model reduces the RMSE score by 0.008, 0.024, and 0.096 compared with those of InstructCV, PromptDiff, and Unified-IO, respectively. On the ADE20K dataset for the Image-to-Seg task, our model achieves RMSE score reductions of 0.044, 0.121, and 0.130 over those of PromptDiff, InstructCV, and Unified-IO, respectively. These results further demonstrate the strong generalization ability of our model when transferred to new datasets, even without task-specific training, highlighting its robustness and adaptability in handling diverse and unseen data scenarios.
\setlength{\tabcolsep}{3.3pt}
\begin{table}[!t]
\renewcommand{\arraystretch}{1.2}
\caption{Comparison of PromptDiff and Our model on Novel Tasks }
\label{table_new_tasks}
\centering
\begin{tabular}{p{2.75cm} p{1cm}<{\centering} p{1cm}<{\centering} p{0.01cm} p{1cm}<{\centering} p{1cm}<{\centering}}
\toprule
\makecell[c]{\multirow{4}{*}{Model}}&\multicolumn{2}{c}{RMSE $\downarrow$}&&\multicolumn{2}{c}{FID $\downarrow$}\\
\cmidrule(r){2-3}
\cmidrule(r){4-6}
 & \multicolumn{1}{c}{{Image-to-}} & \multicolumn{1}{c}{Image-to-} && \multicolumn{1}{c}{Canny-to-} & \multicolumn{1}{c}{Scribble-to-} \\
& \multicolumn{1}{c}{Canny}& \multicolumn{1}{c}{Scribble}&& \multicolumn{1}{c}{Image}& \multicolumn{1}{c}{Image}               \\
                       
\midrule                
PromptDiff\cite{promptdiffusion} &0.345&0.695&&21.72&22.89\\
US-Diffusion (Ours) &\textbf{0.287}&\textbf{0.532}&&\textbf{16.44}&\textbf{16.83}\\
\bottomrule    
\end{tabular}
\end{table}
\subsubsection{\textbf{Comparison on Novel Tasks}}
We further evaluate the generalization ability of our model on novel tasks outside its training scope, including Image-to-Canny, Image-to-Scribble, Canny-to-Image, and Scribble-to-Image tasks. We adopt the canny edge detector\cite{edge} and the PidiNet scribble processor\cite{pidinet} to obtain the ground truth condition maps from the Instruct-Pix2Pix dataset\cite{instructpix2pix} for these tasks. As shown in Table \ref{table_new_tasks}, our model significantly outperforms PromptDiff\cite{promptdiffusion} in generalizing to new tasks, achieving RMSE score reductions of 0.058 on Image-to-Canny task and 0.163 on Image-to-Scribble task, as well as FID score reductions of 5.28 on Canny-to-Image task and 6.06 on Scribble-to-Image task. Notably, on Map2Image tasks, the performance on novel tasks is comparable to that on training tasks.

In summary, our method demonstrates superior generalization to new tasks compared with PromptDiff, particularly for Map2Image tasks. This improvement is driven by the effective integration of SGA and FAL. SGA enhances the model's ability to focus on relevant spatial features, while FAL aligns feature distributions between generated and target images, resulting in more precise and realistic reconstructions. Together, these components enhance the model's robustness and adaptability to unseen tasks, making it a versatile and effective solution for image generation.
\subsection{Inference Efficiency}
The efficient sampling strategy we proposed involves non-uniformly sampling 10 time steps from a total of 1000 time steps, and training and inference are only at these time steps. This is in contrast to the previous approach of training on 1000 time steps but only utilizing 100 time steps during inference.

we compare the inference speed of our model with other models on a single 4090 GPU, focusing on both the number of inference steps and the time required to output one image, as shown in Table \ref{table_inference}. Our US-Diffusion model reduces the number of inference steps to just 1/10 of that required by PromptDiff \cite{promptdiffusion} (10 steps vs 100 steps), leading to a 9.45-fold reduction in inference time (1.58s vs. 14.93s). Compared to the shortest inference time of 4.10s among the comparison methods, our model outputs one image in only 1.58s, achieving a 2.49 times improvement in inference efficiency. Furthermore, subsequent experiments show that the ESS strategy also enhances training efficiency.
\begin{table}[!t]
\renewcommand{\arraystretch}{1.2}
\caption{Comparison of Inference Steps and Inference Time
\label{table_inference}}
\centering
\begin{tabular}{c c c}
\toprule             
 Model &Inference Steps & Inference Time (s)\\       
\midrule                
GLIGEN\cite{gligen}&50 &  17.63 \\
T2I-Adapter\cite{t2i-adapter}&50 &  7.83 \\
Unified-IO\cite{unified-io} &- &  \underline{4.10}\\
InstructCV\cite{instructcv} &100 &  7.59 \\
ControlNet\cite{controlnet} &100 &  18.68 \\
PromptDiff\cite{promptdiffusion} &100 &  14.93 \\
US-Diffusion (Ours) &10 &  \textbf{1.58}\\
\bottomrule               
\end{tabular}
\end{table}

\begin{table*}[htbp]
\renewcommand{\arraystretch}{1.2}
\caption{Ablation Studies of the Design Impact of Each Method. SGA Refers to the Separate $\&$ Gather Adapter; FAL Refers to the Feedback-Aided Learning; ESS Refers to the Efficient Sampling Strategy
\label{table_ablation}}
\centering
\begin{tabular}{c c c c c c p{0.01cm} c c c}
\toprule              
\multirow{3}{*}{SGA}  &\multirow{3}{*}{FAL} &\multirow{3}{*}{ESS}&&Map2Image (FID $\downarrow$)&&&&Image2Map (RMSE $\downarrow$)&\\
\cmidrule(r){4-6}
\cmidrule(r){8-10}
& &  &Seg-to-Image&Hed-to-Image&Depth-to-Image&&Image-to-Seg&Image-to-Hed&Image-to-Depth\\       
\midrule                
& &  &25.32 &23.72 &25.26 &&0.406 &\textbf{0.197} &0.272 \\
\checkmark & & &23.77 &21.77 &22.86 &&0.389 &\underline{0.198} &0.271\\
\checkmark & \checkmark&  &\underline{20.59} &\underline{19.70} &\underline{20.32} &&\underline{0.358} &0.201 &\underline{0.243} \\
\checkmark & \checkmark& \checkmark &\textbf{18.21} &\textbf{16.62} &\textbf{17.05} &&\textbf{0.350} &0.207 &\textbf{0.239} \\
\bottomrule            
\end{tabular}
\end{table*}

\begin{table*}[htbp]
\renewcommand{\arraystretch}{1.2}
\caption{The Results of ControlNet and PromptDiff with and without ESS. Symbol $\dagger$ Refers to the Results Obtained After Training Only for 5000 Steps
\label{table_ESS_controlnet}}
\centering
\begin{tabular}{p{2.5cm} c c c p{0.01cm} c c c }
\toprule              
\makecell[c]{\multirow{3}{*}{Model}} & \multicolumn{3}{c}{Map2Image (FID $\downarrow$)} & & \multicolumn{3}{c}{Image2Map (RMSE $\downarrow$)} \\  
\cmidrule(r){2-4}
\cmidrule(r){6-8}
 & Seg-to-Image & Hed-to-Image & Depth-to-Image && Image-to-Seg & Image-to-Hed & Image-to-Depth \\    
\midrule                
ControlNet (FT)\cite{controlnet} &29.58&\underline{21.35}&25.46&&0.380 &0.240&0.298 \\
ControlNet+ESS$\dagger$  &\underline{27.90}&22.21 &\underline{24.32}&&\underline{0.377}&\underline{0.237} &\underline{0.297} \\
ControlNet+ESS  &\textbf{24.23}&\textbf{20.24}&\textbf{20.49}&&\textbf{0.361}&\textbf{0.226 }&\textbf{0.271}\\
\midrule 
PromptDiff\cite{promptdiffusion} & 25.32 & 23.72 & 25.26 && \underline{0.406} & \textbf{0.196} & \underline{0.272} \\
PromptDiff+ESS$\dagger$  &\underline{22.34}&\underline{21.79} &\underline{23.46}&&0.408&0.227 &0.279 \\
PromptDiff+ESS  &\textbf{21.65}&\textbf{21.19}&\textbf{21.91}&&\textbf{0.395}&\underline{0.214}&\textbf{0.260}\\
\bottomrule           
\end{tabular}
\end{table*}
\subsection{Ablation Studies}
In this section, we conduct ablation studies to evaluate the impact of key designs and hyperparameters on model performance. First, we analyze the contribution of each component design to our model's performance across different tasks. Second, we investigate the effect of varying the weight $\lambda$ of the feedback loss introduced by FAL. Finally, we validate the effectiveness of the proposed ESS on other baseline models, including ControlNet\cite{controlnet} and PromptDiff\cite{promptdiffusion}.
\begin{figure*}[!t]
\centering
\subfloat[]{\includegraphics[width=3.53in]{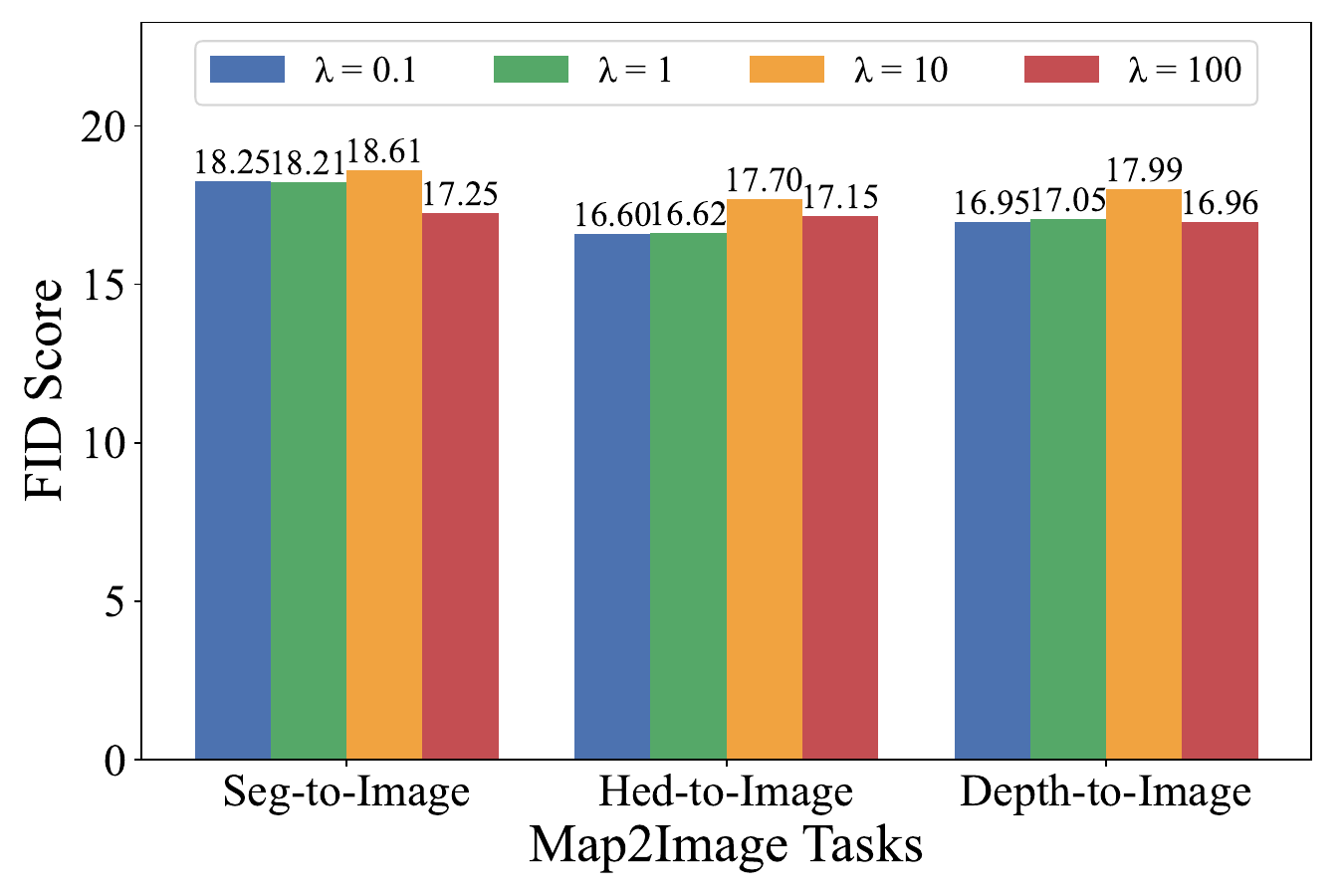}%
\label{fig_first_case}}
\hfil
\subfloat[]{\includegraphics[width=3.53in]{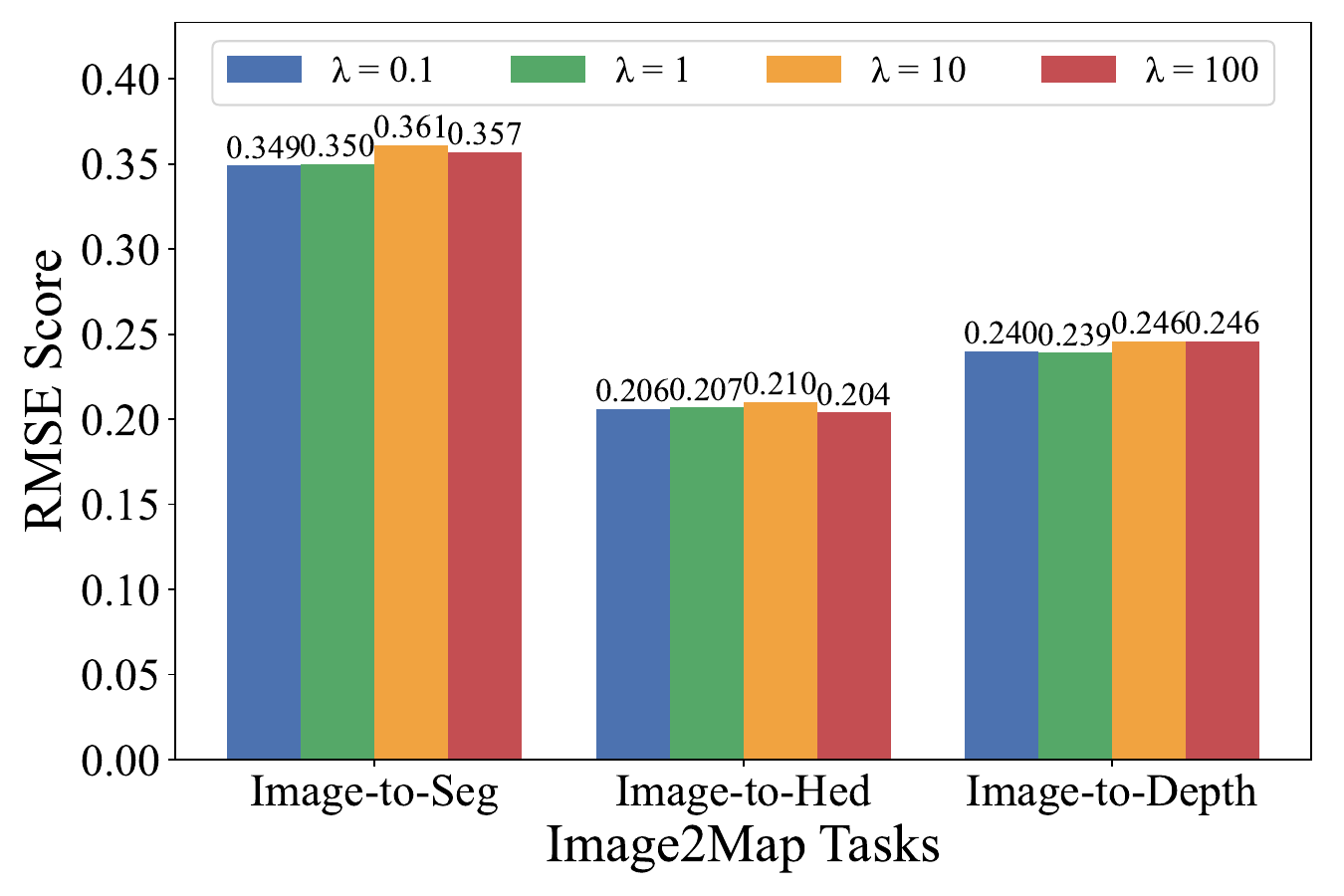}%
\label{fig_second_case}}
\caption{The impact of different $\lambda$ values. (a) The impact of different $\lambda$ values on Image2Map tasks. (b) The impact of different $\lambda$ values on Image2Map tasks.}
\label{fig_sim}
\end{figure*}
\subsubsection{\textbf{Impact of Each Component's Design}}
To verify the impact of each component design on model performance, we conduct ablation experiments, as shown in Table \ref{table_ablation}. The results demonstrate that adding the Separate $\&$ Gather Adapter (SGA) reduces FID score by an average of 1.97 on Map2Image tasks and RMSE score by an average of 0.006 on Image2Map tasks. Further incorporating Feedback-Aided Learning (FAL) leads to additional performance improvements, reducing FID score by an average of 2.60 on Map2Image tasks and RMSE score by an average of 0.019 on Image2Map tasks. Finally, including the Efficient Sampling Strategy (ESS) further enhances performance, reducing FID score by an average of 2.91 on Map2Image tasks and RMSE score by an average of 0.002 on Image2Map tasks. ESS contributes to these gains by focusing training and inference on critical time steps with high-noise levels, ensuring better alignment between training and sampling trajectories, and optimizing resource allocation to the most impactful phases of the denoising process. These results underscore the substantial contribution of each component to the overall performance of the model.
\begin{figure*}[htbp]
  \centering
  \includegraphics[scale=0.918]{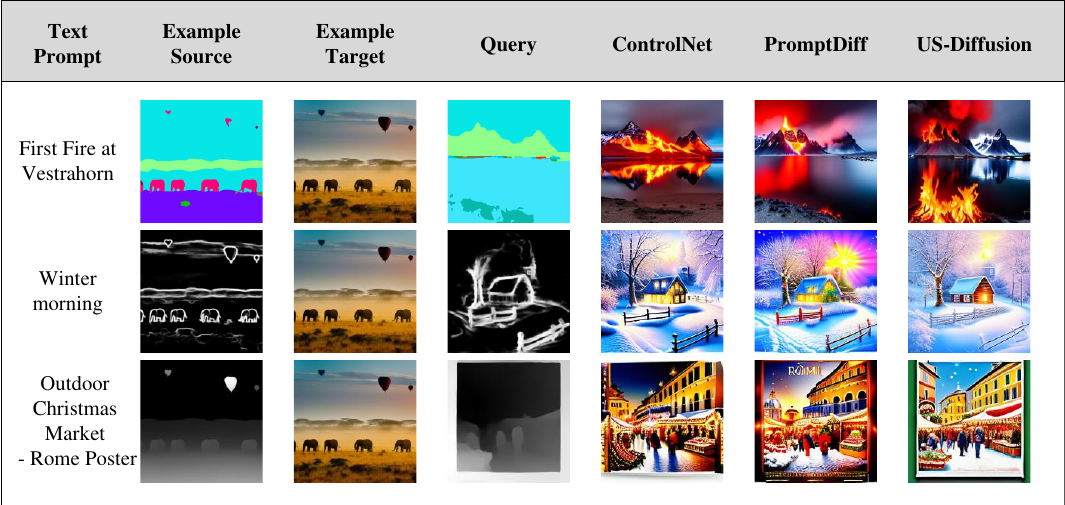}
  \caption{Comparison of PromptDiff\cite{promptdiffusion} and our US-Diffusion model for Map2Image tasks. From this figure, it is evident that the controllable generation results of our US-Diffusion model align more accurately with the spatial details specified by the input query condition-map images, particularly in terms of object placement and structure.}
  \label{fig_map2image}
\end{figure*}
\begin{figure*}[htbp]
  \centering
  \includegraphics[scale=0.918]{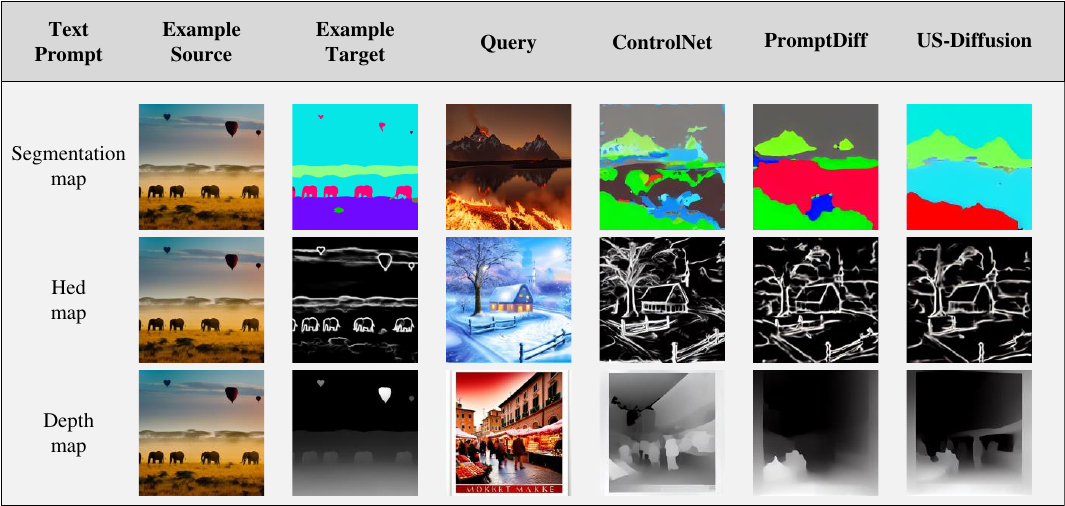}
  \caption{Comparison of map prediction results for Image2Map tasks. The figure demonstrates that the US-Diffusion map prediction results align with the overall layout, objects, structures, and contours in the predicted map when compared to the input query images.}
  \label{fig_image2map}
\end{figure*}
\subsubsection{\textbf{Impact of Different $\lambda$ Values}}
We analyze the influence of varying weights $\lambda$ for the feedback loss in FAL, as illustrated in Fig. \ref{fig_sim}. Both Fig. \ref{fig_sim}(a) and \ref{fig_sim}(b) suggest that different $\lambda$ values have a certain impact on performance. Smaller $\lambda=0.1$ and $\lambda = 1$ achieves better balanced results across both Map2Image and Image2Map tasks by providing moderate feedback signals. Conversely, a larger $\lambda$ (e.g., $\lambda = 10$ or 100) overemphasizes the feedback signals, which distorts the learning process and degrades performance. For example, when $\lambda$ is set to 10, both the FID and RMSE scores increase significantly, indicating a notable decline in the performance of both Map2Image and Image2Map tasks under this setting. This demonstrates a trade-off between feedback strength and task-specific optimization. For simplicity, we adopt $\lambda = 1$ as the default configuration.

\subsubsection{\textbf{Versatility of ESS on other baselines}}
We further verify the versatility of the proposed Efficient Sampling Strategy (ESS) on ControlNet\cite{controlnet} and PromptDiff\cite{promptdiffusion}, as shown in Table \ref{table_ESS_controlnet}. 

From the upper part of this table, we observe that ControlNet retrained with ESS outperforms the original without ESS under identical training conditions. For Map2Image tasks, ESS reduces the FID score by an average of 3.81, with reductions up to 5.35 and as low as 1.11. For Image2Map tasks, ESS reduces the RMSE score by an average of 0.020, with the highest reduction of 0.027 and the lowest of 0.014. Additionally, ControlNet with ESS, trained for only 5,000 steps, achieves performance comparable to the original ControlNet trained for 10,000 steps. This demonstrates that ESS enhances training efficiency and reduces resource consumption. 

From the bottom part of the table, ESS significantly improves the performance of PromptDiff across most tasks. For Map2Image tasks, ESS reduces the FID score by an average of 3.18, with the highest reduction of 3.67 and the lowest of 2.53. Even with only 5,000 training steps, the model outperforms the original PromptDiff trained for 10,000 steps. For Image2Map tasks, ESS improves performance on two of three subtasks: Image-to-Seg task and Image-to-Depth task. PromptDiff with ESS also achieves comparable results with half the training steps, further demonstrating its efficiency in accelerating convergence. However, Image-to-Hed shows slight degradation, suggesting task-specific sensitivity.

These results indicate that ESS not only increases training efficiency but also enhances model performance, making it a versatile solution for diffusion-based frameworks. While the slight trade-off on Image-to-Hed  requires further study, it does not diminish ESS’s overall effectiveness in multi-task training.

\subsection{Qualitative Results}
In this section, we provide a qualitative comparison of the results of our model with those of other methods on the training tasks. 
The qualitative comparison of the controllable generation results among our model, the ControlNet\cite{controlnet}, and PromptDiff\cite{promptdiffusion} is presented in Fig. \ref{fig_map2image}. This figure demonstrates that the controllable generation results of our model align more closely with the spatial details provided by the condition-map, particularly in terms of object and structure. This also indirectly confirms that our method enables the model to more effectively learn the underlying semantic information. For example, for the Seg-to-Image task in the first row, ControlNet generates an image where the area that should be water in the query image is incorrectly rendered as land. In the image generated by PromptDiff, two unreasonable black blocks appear on the mountain's peak on the left, and the water surface fails to reflect the mountain's reflection. In contrast, the image generated by our model presents a more reasonable depiction of object shapes and layout. The qualitative comparison of map prediction results between our model, the ControlNet model, and PromptDiff is shown in Fig. \ref{fig_image2map}. It is evident from this figure that our map prediction results align with the overall layout, objects, structures, and contours in the predicted map in comparison to the input image. For example, for the Image-to-Seg task in the first row, the generated segmentation map of the mountain better aligns with the shape and layout of the mountain in the query image.

\section{Conclusion}
This paper has presented US-Diffusion, a novel framework that boosts in-context learning of diffusion models by addressing challenges like limited underlying semantics and low computational efficiency in image generation tasks. Our contributions include the introduction of the Separate \& Gather Adapter (SGA), Feedback-Aided Learning (FAL), and Efficient Sampling Strategy (ESS), each designed to boost in-context learning, underlying semantics capture, and inference speed. Experimental results demonstrate that US-Diffusion outperforms the state-of-the-art method, achieving approximately a 9.45-fold increase in inference speed, substantial improvements in image generation quality and robust generalization to new datasets and tasks. These advancements make US-Diffusion a highly efficient and adaptable solution for real-time applications in diverse computer vision tasks. In the future, we will focus on extending these improvements to other generative models and further enhancing the model's generalization capabilities.
\bibliographystyle{IEEEtran}
\bibliography{IEEEabrv,reference}

\begin{IEEEbiography}[{\includegraphics[width=1in,height=1.25in,clip,keepaspectratio]{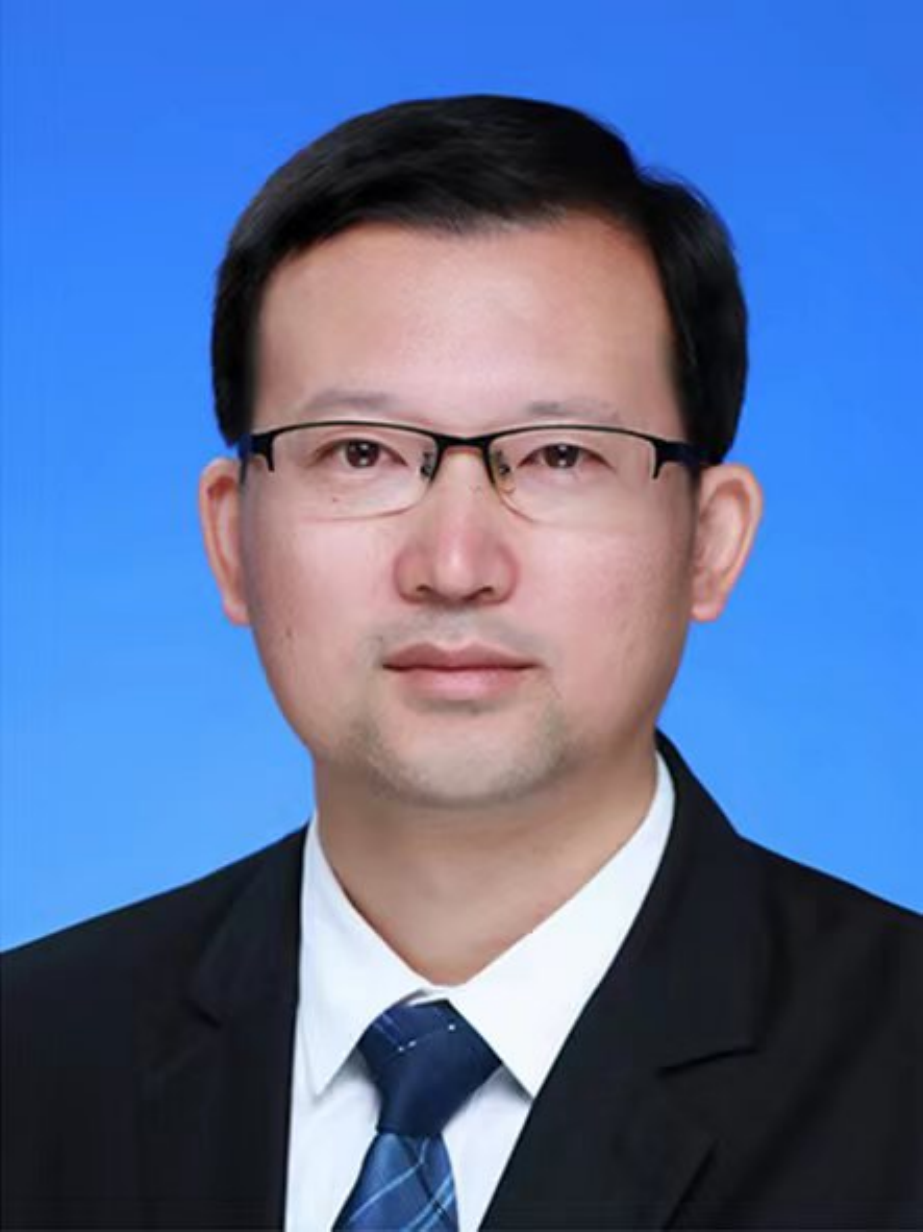}}]{Zhong Ji} (Senior Member, IEEE)
received the Ph.D. degree in signal and information processing from Tianjin University, Tianjin, China, in 2008. He is currently a Professor with the School of Electrical and Information Engineering, Tianjin University. He has authored over 100 technical articles in refereed journals and proceedings. His current research interests include multimedia understanding, zero/few-shot learning, and cross-modal analysis.
\end{IEEEbiography}

\begin{IEEEbiography}[{\includegraphics[width=1in,height=1.25in,clip,keepaspectratio]{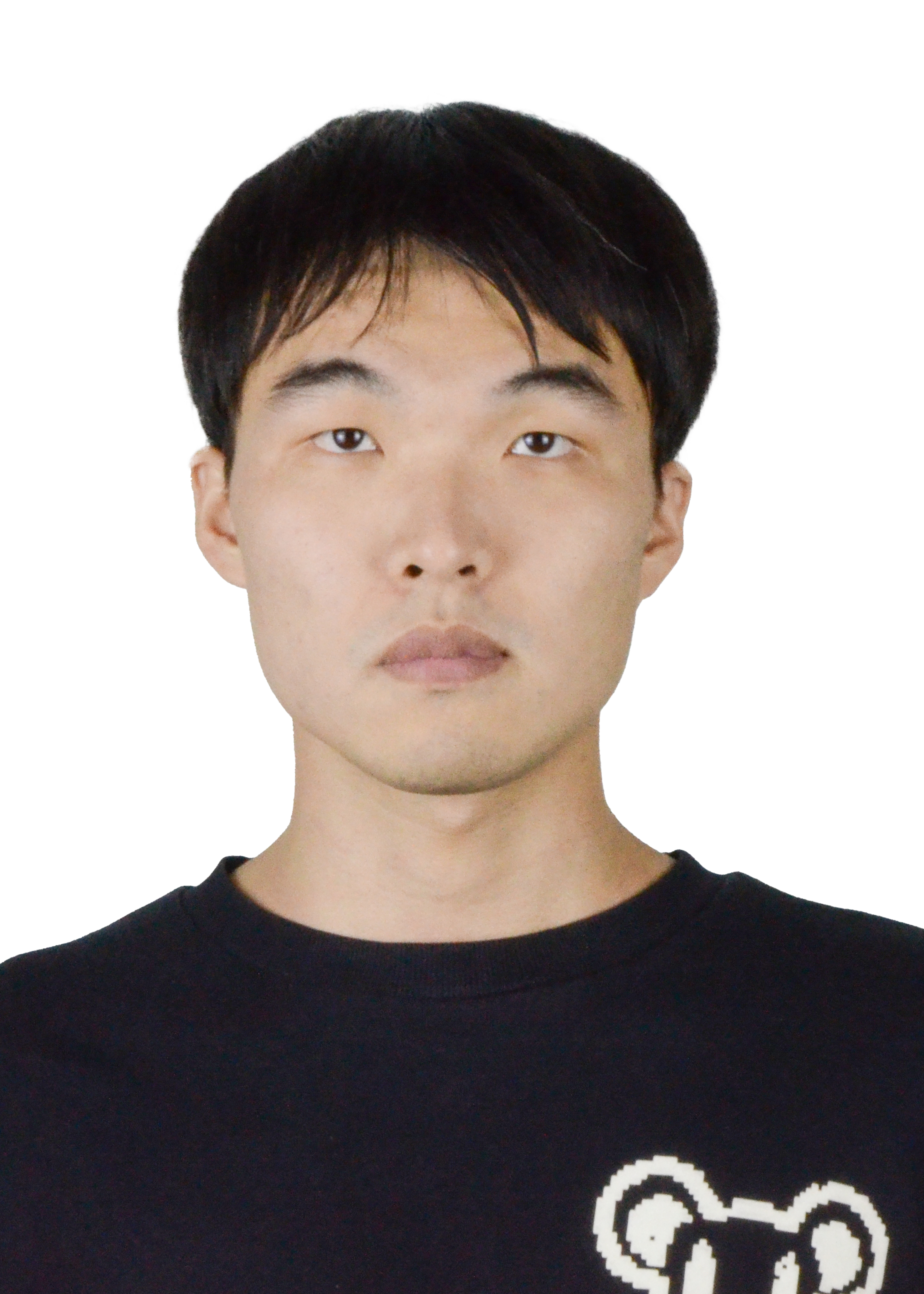}}]{Weilong Cao}
received the B.S. degree in communication engineering from Tianjin University, Tianjin, China, in 2023. He is currently pursuing the M.S. degree in the School of Electrical and Information Engineering, Tianjin University. His current research interests include multi-task vision foundation models and computer vision.
\end{IEEEbiography}

\begin{IEEEbiography}[{\includegraphics[width=1in,height=1.25in,clip,keepaspectratio]{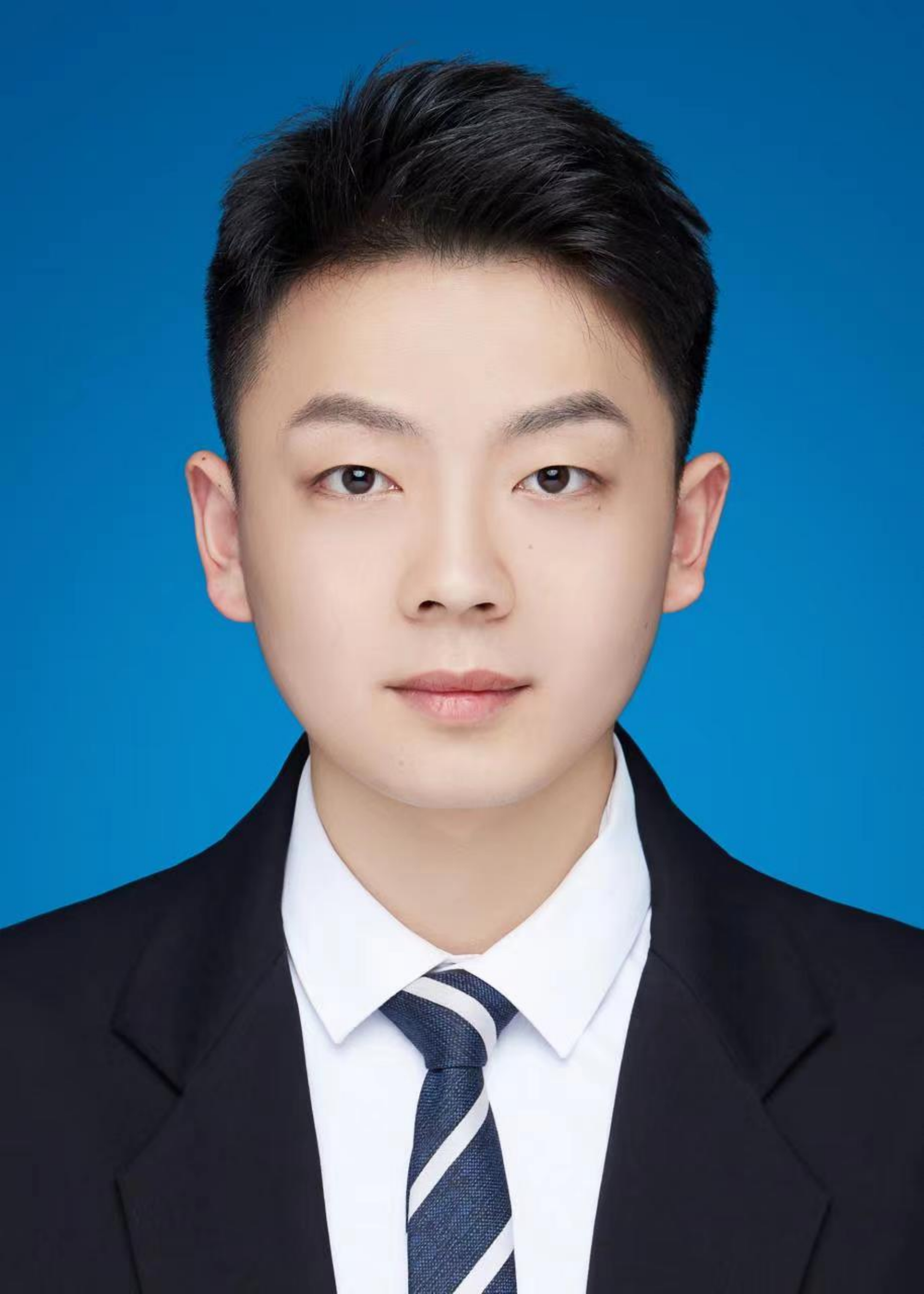}}]{Yan Zhang}
received the M.S. degree in control engineering from Tianjin University, Tianjin, China, in 2021. He is currently pursuing a Ph.D. degree in the School of Electrical and Information Engineering, Tianjin University. His current research interests include multi-modal learning and computer vision.
\end{IEEEbiography}

\begin{IEEEbiography}[{\includegraphics[width=1in,height=1.25in,clip,keepaspectratio]{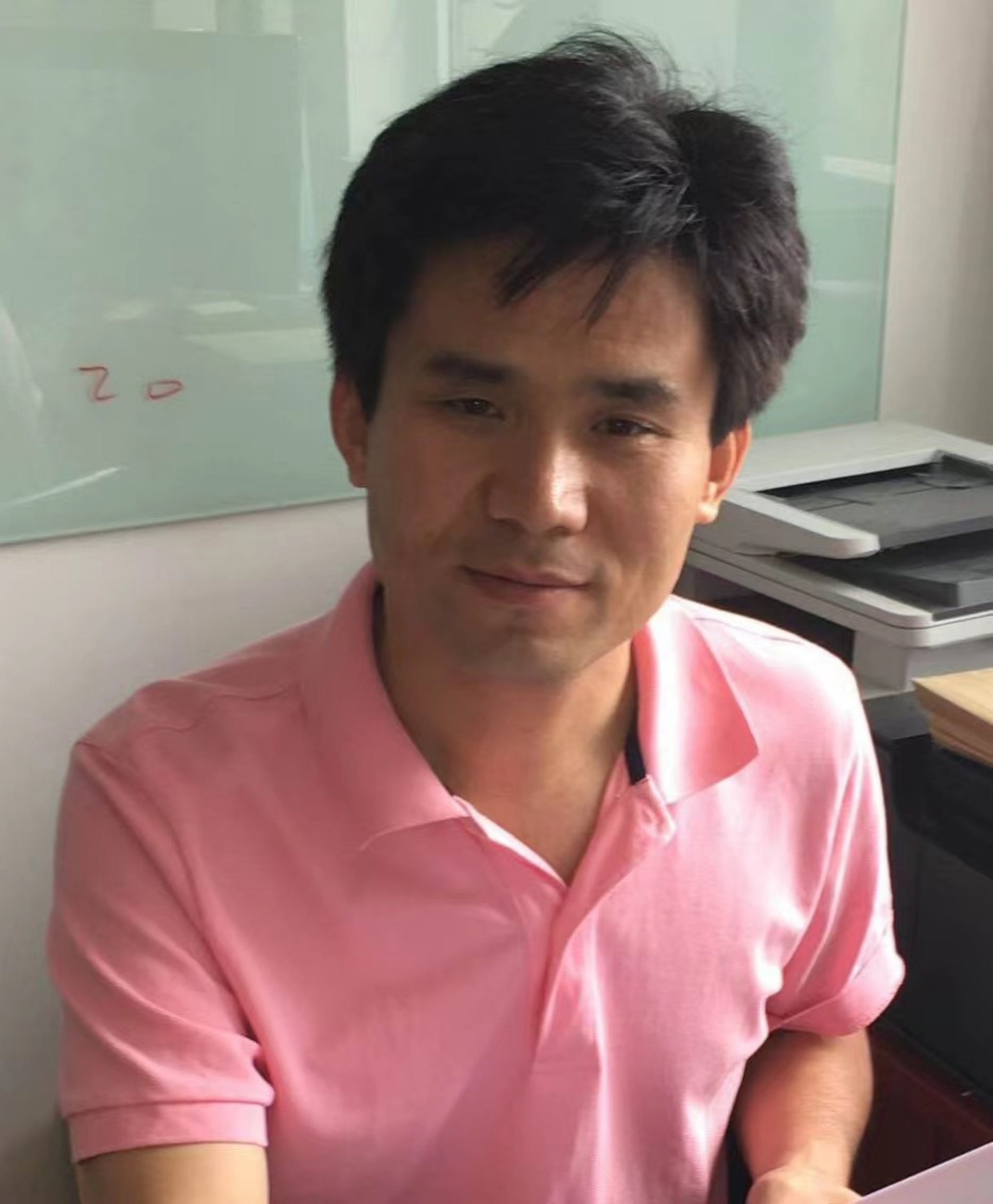}}]{Yanwei Pang}(Senior Member, IEEE)
received the Ph.D. degree in electronic engineering from the University of Science and Technology of China, Hefei, China, in 2004. He is currently a Professor with School of Electrical and Information Engineering, Tianjin University, Tianjin, China. He has authored over 150 technical articles in refereed journals and proceedings. His current research interests include object detection and recognition, vision in bad weather, and computer vision.
\end{IEEEbiography}

\begin{IEEEbiography}[{\includegraphics[width=1in,height=1.25in,clip,keepaspectratio]{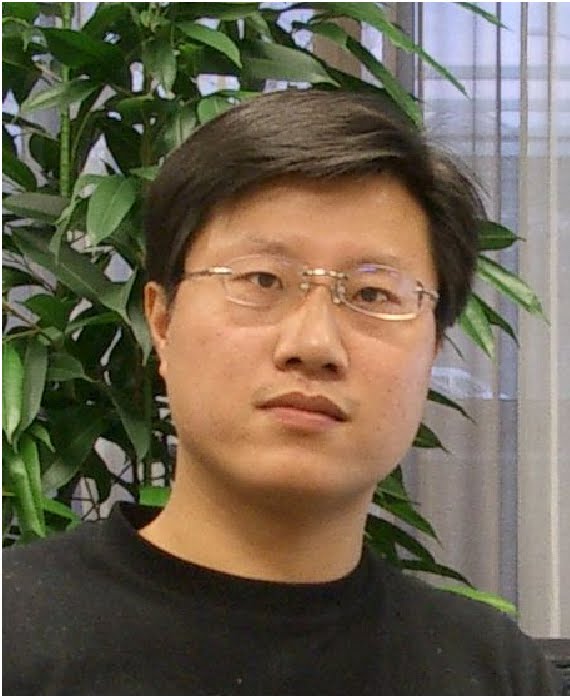}}]{Jungong Han}(Senior Member, IEEE)
received the Ph.D. degree in telecommunication and information system from Xidian University, Xi'an, China, in 2004. He is  a Tenured Professor with the Department of Automation, Tsinghua University. He has published over 200 articles, including more than 80 IEEE Transactions and more than 50A* conference articles. His research interests span the fields of video analysis, computer vision, and applied machine learning. He is a Fellow of the International Association of Pattern Recognition.
\end{IEEEbiography}

\begin{IEEEbiographynophoto}{Xuelong Li}(Fellow, IEEE)
is the Chief Technology Officer (CTO) and the Chief Scientist of the Institute of Artificial Intelligence (TeleAI) of China Telecom. 
\end{IEEEbiographynophoto}

\vspace{11pt}
\end{document}